%% file: neurips_2025.tex
\documentclass{article}

    \PassOptionsToPackage{numbers, compress}{natbib}

\usepackage[preprint]{neurips_2025}

\usepackage[utf8]{inputenc} 
\usepackage[T1]{fontenc}    
\usepackage{hyperref}       
\usepackage{url}            
\usepackage{booktabs}       
\usepackage{subcaption}    
\usepackage{amsfonts}       
\usepackage{nicefrac}       
\usepackage{microtype}      
\usepackage{amssymb}
\usepackage{amsmath}
\usepackage{wasysym}  
\usepackage{graphicx}
\usepackage{longtable}
\usepackage{multirow}
\usepackage{geometry}
\usepackage{array}      
\usepackage{bm}
\usepackage[leftcaption]{sidecap}
\usepackage{colortbl}
\usepackage{adjustbox}
\usepackage[table]{xcolor}
\usepackage{wrapfig}
\definecolor{lightorange}{RGB}{255, 230, 204} 
\definecolor{moderateorange}{RGB}{255, 178, 102} 

\newcolumntype{P}[1]{>{\raggedright\arraybackslash}p{#1}}

\title{
TerraFM: A Scalable Foundation Model for Unified Multisensor Earth Observation}

\author{
\fontsize{10}{10}\selectfont
Muhammad Sohail Danish$^{1}$ \quad
Muhammad Akhtar Munir$^{1}$ \quad
Syed Roshaan Ali Shah$^{2}$ \\ \quad
\textbf{Muhammad Haris Khan}$^{1}$ \quad
\textbf{Rao Muhammad Anwer}$^{1,3}$ \quad
\textbf{Jorma Laaksonen}$^{3}$ \\ \quad
\textbf{Fahad Shahbaz Khan}$^{1,4}$ \quad
\textbf{Salman Khan}$^{1,5}$ \\
$^1$Mohamed bin Zayed University of Artificial Intelligence \quad
$^2$University College London \\
$^3$Aalto University \quad
$^4$Link\"oping University, Sweden \quad
$^5$Australian National University \quad
}

\begin{document}

\maketitle
\input{sec/01_abstract}

\input{sec/02_introduction}

\input{sec/02_related_work}

\input{sec/03_methodology}

\input{sec/02_t_dataset}

\input{sec/04_experiments}

\section{Conclusion}
In this work, we introduced TerraFM, a unified and scalable foundation model (FM) specifically designed for multisensor EO. 
Given the unique nature of EO data, our approach pays special treatment to sensor heterogeneity, scale-invariance, and class-frequency imbalance which is critical for building generalizable EO FMs.
Our pretraining approach leverages contrastive learning to  obtain geographically and spectrally aware representations from large-scale Sentinel-1 and 2 data.
Specifically, we integrate  modality-specific patch embeddings, adaptive cross-attention fusion, and a dual-centering contrastive learning objective to enrich the representations on heterogeneous RS data. 
Our extensive evaluations on GEO-Bench and Copernicus-Bench demonstrate that TerraFM consistently outperforms SoTA self-supervised ViT models across both classification and  segmentation tasks.

{\small
\bibliographystyle{chicago}
\bibliography{natbib}
}







\newpage
\appendix

\renewcommand{\thefigure}{A\arabic{figure}}
\setcounter{figure}{0}  
\renewcommand{\thetable}{A\arabic{table}}
\setcounter{table}{0}

\newcounter{secnumber}
\setcounter{secnumber}{1} 
\renewcommand{\thesection}{S\arabic{secnumber}}

{\LARGE \textbf{Supplementary Material}}

\input{sec/05_supp}

\end{document}

%% file: sec/01_abstract.tex
\begin{abstract}

Modern Earth observation (EO) increasingly leverages deep learning to harness the scale and diversity of satellite imagery across sensors and regions.
While recent foundation models have demonstrated promising generalization across EO tasks, many remain limited by the scale, geographical coverage, and spectral diversity of their training data, factors critical for learning globally transferable representations. In this work, we introduce \textbf{TerraFM}, a scalable self-supervised learning model that leverages globally distributed Sentinel-1 and Sentinel-2 imagery, combined with large spatial tiles and land-cover aware sampling to enrich spatial and semantic coverage. By treating sensing modalities as natural augmentations in our self-supervised approach, we unify radar and optical inputs via modality-specific patch embeddings and adaptive cross-attention fusion. 
Our training strategy integrates local-global contrastive learning and introduces a dual-centering mechanism 
that incorporates class-frequency-aware regularization to address long-tailed distributions in land cover.
TerraFM achieves strong generalization on both classification and segmentation tasks, outperforming prior models on GEO-Bench and Copernicus-Bench. 
Our code and pretrained models are publicly available at \href{https://github.com/mbzuai-oryx/TerraFM}{https://github.com/mbzuai-oryx/TerraFM}.
\end{abstract}

%% file: sec/02_introduction.tex
\section{Introduction}
EO provides systematic measurements of the surface of the earth, supporting a wide spectrum of critical applications such as land use monitoring~\cite{wang2023review}, crop evaluation~\cite{prodhan2021deep, kussul2017deep}, urban development~\cite{yu2023urban}, and disaster response~\cite{huot2022next, sarkar2023sam, rahnemoonfar2021floodnet}. 
These capabilities are enabled by a growing fleet of earth-observing satellites, most notably the Sentinel missions, which deliver multi-modal, multi-temporal data at a global scale~\cite{torres2012gmes, drusch2012sentinel}. The rise of deep learning, particularly deep neural networks (DNNs), has fundamentally reshaped how EO data is processed and interpreted~\cite{wang2025towards, szwarcman2024prithvi, tseng2025galileo}. 
Modern DNNs enable automated extraction of spatial and semantic patterns from raw imagery, driving downstream tasks such as scene classification, object detection, and semantic segmentation~\cite{astruc2024anysat, waldmann2025panopticon, kuckreja2024geochat, tseng2025galileo, fuller2023croma}. 
These models offer a scalable and adaptive alternative to traditional hand-engineered pipelines by learning generalizable representations directly from the data~\cite{guo2024skysense}. As EO datasets continue to expand in scale, diversity, and complexity, DNNs have become the foundation for building high-capacity models capable of generalizing across geographies, modalities, and tasks~\cite{szwarcman2024prithvi, astruc2024anysat, waldmann2025panopticon}.

Remote sensing data is inherently multimodal, comprising diverse sensor types such as optical, SAR, and multispectral imagery. 
Traditional EO pipelines often focus on single-modality inputs, typically high-resolution optical imagery, limiting the model's ability to generalize across varying sensing conditions. 
In contrast, multimodal and multispectral data sources, such as Sentinel-1 SAR and Sentinel-2 Level-1C/Level-2A optical bands, capture complementary structural and spectral information, enabling richer scene understanding \cite{han2024bridging, fuller2023croma}. 
Foundation models that embrace this diversity have demonstrated superior transferability across tasks and geographies \cite{tseng2025galileo, guo2024skysense}. 
However, variation in ground sampling distance (GSD) across EO data makes tile size a critical factor; smaller tiles capture local detail but risk overfitting to texture, while larger tiles provide broader semantic context but require scale-robust architectures~\cite{reed2023scale}.
Recent works like AnySat and msGFM \cite{han2024bridging, astruc2024anysat} have shown that scale-invariant modeling and mixed-resolution pretraining lead to more robust and generalizable representations. 
Crucially, large-scale sampling across geographies and resolutions enables EO foundation models to learn invariant features across sensors and global conditions.

\begin{SCfigure}[][t]   
  \caption{Performance comparison across GEO-Bench classification tasks using supervised and kNN-based evaluation protocols. Specifically, supervised fine-tuning assesses the adaptability of the learned base representations and the frozen-encoder with kNN head assesses the generalization of representations.
  Five recent EO foundation models (AnySat \cite{astruc2024anysat}, DOFA \cite{xiong2024neural}, Galileo \cite{tseng2025galileo}, Prithvi-2.0 \cite{szwarcman2024prithvi}, Satlas \cite{satlaspretrainlargescaledatasetremote}) performances are shown compared to our proposed TerraFM.
  TerraFM consistently outperforms existing models across modalities and evaluation settings, showing strong generalization.}
  \centering
  \includegraphics[width=0.6\textwidth]{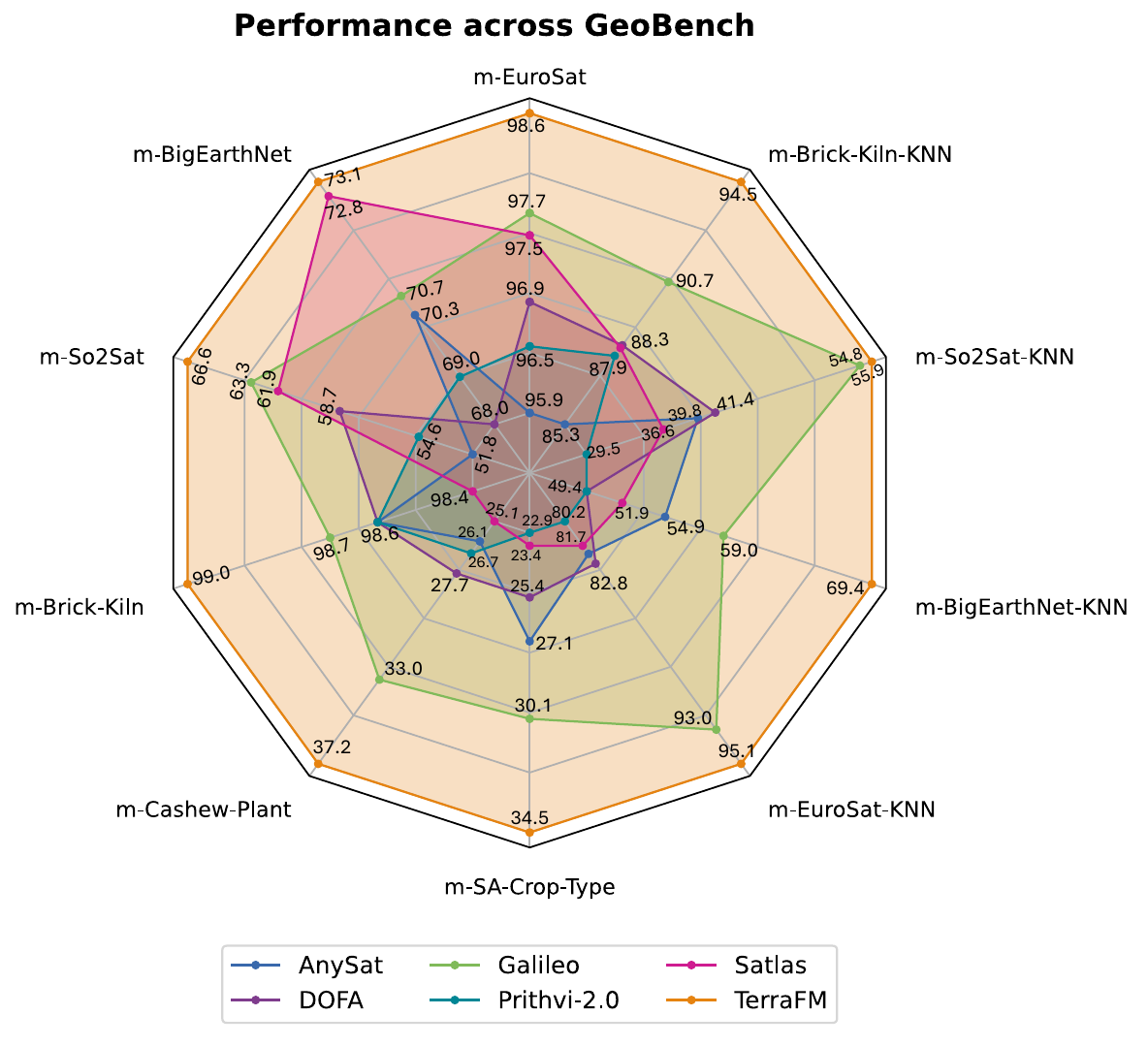}
  \label{fig:spider_gb}
\end{SCfigure}

As EO foundation models scale to accommodate diverse sensor inputs and resolutions, two dominant pretraining paradigms have emerged: masked autoencoders (MAE) and contrastive learning. 
Although MAEs focus on reconstructing the spatial structure, their reliance on RGB-centric ViTs limits their adaptability to multispectral or SAR inputs with varying spectral dimensions~\cite{li2024s2mae, szwarcman2024prithvi}. 
In contrast, contrastive approaches such as DINO \cite{caron2021emerging, oquab2023dinov2} and its adaptations to remote sensing \cite{tseng2025galileo, fuller2023croma, waldmann2025panopticon} offer modality-agnostic training by aligning global and local views through student-teacher distillation. 
However, the expansive spatial coverage of EO datasets introduces new challenges: large portions of satellite imagery are semantically sparse or uninformative, and naïve sampling can lead to representation bias. 
This requires intelligent sampling that prioritizes semantically diverse regions, guided by land cover priors, for balanced and efficient representation learning.

To address these limitations in standard ViTs, particularly their RGB-centric design, lack of modality awareness, and unimodal self-supervision, we introduce \textbf{TerraFM}, a unified foundation model tailored for remote sensing.
First, we propose a \textit{Modality-Specific Patch Embedding} module, which replaces the shared projection in standard ViTs with modality-aware embeddings adapted to multispectral and SAR data. 
This enables flexible handling of sensor-specific spectral profiles while preserving spatial structure. 
To enhance scale-invariance and cross-view consistency, we adopt multi-crop learning within a self-supervised teacher-student framework, promoting robust representation learning through global-local alignment. 
Further, we interpret different aligned modalities (S1-SAR, S2-L1C, S2-L2A) as complementary views of the same scene and introduce a \textit{Cross-Attention Fusion} module that dynamically aggregates modality-specific tokens using learnable spatial queries. 
This allows the model to selectively emphasize sensor contributions at each spatial location. 
Finally, to mitigate long-tailed land cover distribution issues prevalent in EO data, we introduce a \textit{Dual Centering} mechanism into the distillation process. 
This leverages WorldCover \cite{zanaga2022esa} derived class statistics to compute a frequency-aware center, improving balance across dominant and rare semantic categories without requiring supervised objectives. 
Our key contributions are as follows.

\noindent\textbf{Contributions:} \textbf{(1)} A \textit{modality-specific patch embedding} mechanism is introduced to generalize ViTs across heterogeneous remote sensing modalities with varying spectral dimensions. \textbf{(2)} We treat sensor modalities as natural augmentations and introduce a \textit{cross-attention fusion} block that unifies multi-modal inputs within a shared encoder. \textbf{(3)} To address long-tailed LULC distributions, a \textit{dual-centering} strategy is incorporated to regularize representation learning using class-frequency-aware statistics. \textbf{(4)} Extensive experiments on \textit{GEO-Bench} and \textit{Copernicus-Bench} demonstrate leading performance across multiple downstream tasks using globally distributed data (Fig.~\ref{fig:spider_gb}).

%% file: sec/02_related_work.tex
\section{Related Work}

\noindent\textbf{Self-supervised Pretraining:}
MAEs~\cite{he2022masked} have become a popular choice for self-supervised pretraining in remote sensing by reconstructing masked image regions using ViT~\cite{dosovitskiy2021an}. 
Variants like Scale-MAE~\cite{reed2023scale} and MC-MAE~\cite{gao2022mcmae} enhance robustness across spatial scales via scale-aware encodings and convolutional tokenizers. 
However, MAEs struggle to scale to multisensor EO data, as their RGB-centric tokenization and reconstruction objectives limit generalization to multispectral and SAR modalities with diverse channel structures~\cite{xie2023data, li2024s2mae}.

Unlike MAEs, self-supervised contrastive learning focuses on learning discriminative representations by comparing semantically similar and dissimilar views. 
Remote sensing approaches~\cite{tang2023cross, fuller2023croma, waldmann2025panopticon} leverage spatial and spectral augmentations to create diverse yet consistent views. 
CROMA~\cite{fuller2023croma} combines contrastive and masked autoencoding losses, while Cross-Scale MAE~\cite{tang2023cross} blends generative and contrastive objectives for multi-scale learning. 
Student-teacher frameworks like DINO~\cite{caron2021emerging, oquab2023dinov2} scale contrastive learning via EMA-updated teachers and global-local view alignment with centering to prevent collapse. 
These strategies are well-suited for EO, where multimodal imagery can act as natural augmentations, enabling scalable, label-free training and broad generalization.

\begin{table}[t]
\centering
\small
\caption{Comparison of recent remote sensing foundation models across modalities, scale, and benchmarks. \textbf{TerraFM} uniquely blends large tile size (534), WorldCover-informed metadata, and global-scale training (18.7M samples) with evaluation on both GEO-Bench and Copernicus-Bench.}
\label{tab:related}
\resizebox{\linewidth}{!}{%
\begin{tabular}{P{2.2cm}P{3.6cm}P{1.6cm}P{1.6cm}P{1.5cm}P{1.3cm}P{2.6cm}P{1.5cm}}
\toprule
\textbf{Model} & \textbf{Modalities} & \textbf{Scale} & \textbf{Resolution} & \textbf{TileSize} & \textbf{Metadata} & \textbf{Benchmarks} & \textbf{Pixels (\textasciitilde T)} \\
\midrule \midrule
SatMAE++      & S2, RGB                                                  & $\sim$1.2 M    & 10–60 m     & 224, 96   & No                        & 6 DS               & 0.12 \\
Galileo       & S1, S2, NDVI, ESA WC etc                                 & $\sim$3–10.9 M & 10 m        & 96 (flex) & Yes                       & GEO + 5 DS         & 1.58 \\
CROMA         & S1, S2                                                   & $\sim$1 M      & 10 m        & 96, 120   & No                        & 7 DS               & 0.98 \\
SoftCon       & S1, S2                                                   & $\sim$0.78 M   & 10 m        & 224       & Yes                       & 4 GEO + 7 DS       & 0.76 \\
AnySat        & Aerial, S1/S2, MODIS, etc.                                & 11.1 M         & 0.2–250 m   & 10–240    & No                        & 11 DS              & 0.17 \\
Prithvi-2     & S2, HLS                                                  & 4.2 M          & 30 m        & 224       & Yes                       & GEO + 9 SME        & 5.06 \\
DOFA          & S1, S2, EnMAP, etc.                                      & $\sim$8 M      & 1–30 m      & 512, 128  & Yes                       & GEO + 2 DS         & 6.74 \\
Panopticon    & S1, S2, WV2/3, NAIP                                      & $\sim$2.6 M    & 0.3–100 m   & 96, 224   & Yes                       & GEO + 10 DS        & 2.34 \\
MMEarth       & S1, S2, DEM, etc.                                        & $\sim$7.2 M    & 0.3–100 m   & 128       & Yes                       & 5 GEO              & 0.51 \\
msGFM         & RGB, S2, SAR, DSM                                        & $\sim$2 M      & 0.1–30 m    & 192       & No                        & 5 DS               & 0.44 \\
Copernicus-FM & S1–S5P, DEM                                              & 18.7 M         & 10 m–1 km   & Mixed     & Yes                       & Cop-Bench          & 5.12 \\ \midrule
\textbf{TerraFM} (Ours) & S1, S2 L1C/L2A                                 & 18.7 M         & 10–60 m     & 534       & Yes                       & GEO + Cop-Bench    & \textbf{23.32} \\
\bottomrule
\end{tabular}}
\end{table}

\noindent\textbf{Remote Sensing FMs:}
Recent advances in remote sensing foundation models (FMs) have scaled self-supervised learning across architecture types, modalities, training sizes, tile resolutions, and metadata usage (Tab.~\ref{tab:related}). 
Multimodal integration is central to recent FMs like \cite{guo2024skysense, wang2025towards, waldmann2025panopticon, astruc2024anysat, tseng2025galileo, han2024bridging}. 
SkySense \cite{guo2024skysense} applies contrastive learning to temporal-multimodal data but requires large-scale compute. 
CopernicusFM~\cite{wang2025towards} fuses Sentinel modalities via metadata-aware networks but faces scaling issues with heterogeneous inputs. 
Panopticon~\cite{waldmann2025panopticon} and AnySat~\cite{astruc2024anysat} align cross-modal views through contrastive training, while Galileo~\cite{tseng2025galileo} uses shared embeddings for SAR and multispectral fusion. 
Fus-MAE~\cite{chan2024fus} adopts attention-based fusion without contrastive loss, limiting generalization.

Prithvi-2~\cite{szwarcman2024prithvi} is restricted to single-modal optical data with temporal-spatial modeling. 
DOFA~\cite{xiong2024neural}, msGFM~\cite{han2024bridging}, and AnySat~\cite{astruc2024anysat} address resolution variability using mixed tile sizes or scale-adaptive designs. 
Our 534px tiles capture broader spatial context than prior RSFMs. 
While CopernicusFM~\cite{wang2025towards} and DOFA~\cite{xiong2024neural} incorporate metadata, we leverage land cover (LULC) priors for semantically informed learning. 
Both CopernicusFM and our model are trained on 18.7M  samples, but ours uses over 23T pixels during pretraining, scaling the 5.1T used by Copernicus-Pretrain~\cite{wang2025towards} over 4x.

%% file: sec/03_methodology.tex
\begin{figure}[t]
  \centering
    \includegraphics[width=\linewidth]{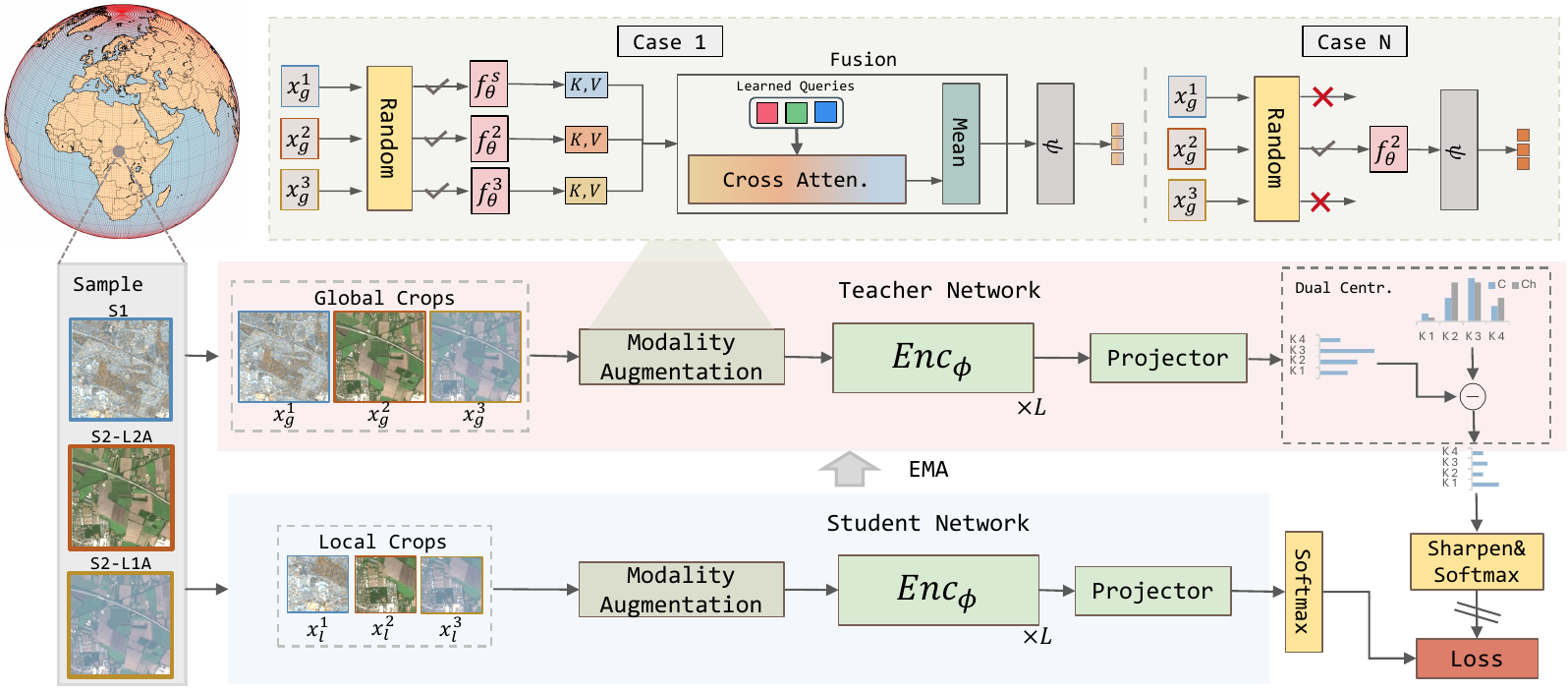}
    \caption{Overall architecture of \textbf{TerraFM}. 
   It unifies student-teacher contrastive framework with modality augmentation with cross-attention fusion, and a new dual centering regularization. \textbf{TerraFM} is founded on ViT backbone and is trained on 18.7M globally distributed samples for pre-training and utilizes large-tile inputs for encoding broader spatial context. For illustration, RGB channels from S2-L2A and S2-L1C are selected, and S1 is visualized using a false-color RGB composite.}
    \label{fig:architecture}
\vspace{-1em}
\end{figure}

\section{TerraFM: A Scalable Multisensor Foundational Model}
Unlike prior remote sensing foundation models, our approach integrates a student–teacher contrastive learning framework with dual centering (to balance long-tailed classes), modality-as-augmentation (to learn cross-modal invariances), and cross-attention fusion (to aggregate multi-sensor context), as illustrated in Fig.~\ref{fig:architecture}. 
Built on a ViT backbone and trained on 18.7M globally distributed samples using 534$\times$534 tiles, \textbf{TerraFM} captures broader spatial context and generalizes effectively across sensing modalities and geographies, achieving strong results on diverse downstream benchmarks.

\subsection{Architecture}
\label{sec:preliminaries}

We use globally distributed remote sensing imagery organized over a spatial grid, partitioning the earth's surface into fixed-size tiles \cite{francis2024major} (e.g., $5.34\,\text{km} \times 5.34\,\text{km}$). Each spatial unit, denoted as \( s \), represents one such grid cell. For each sample, we observe a set of co-registered EO modalities:
\[
\mathcal{M} = \{ \text{S1},\; \text{S2-L1C},\; \text{S2-L2A} \}, 
\]
where \textbf{S1} corresponds to Sentinel-1 SAR (Synthetic Aperture Radar), and \textbf{S2-L1C} and \textbf{S2-L2A} represent two processing levels of Sentinel-2 optical imagery: Level-1C (top-of-atmosphere reflectance) and Level-2A (bottom-of-atmosphere surface reflectance), respectively.
Each modality \( m \in \mathcal{M} \) provides a multi-channel image \( \bm{x}^{m} \in \mathbb{R}^{H \times W \times C_m} \), where \( H \) and \( W \) denote spatial dimensions, and \( C_m \) is the number of spectral channels for modality \( m \). For example, Sentinel-1 contains two channels (VV and VH polarizations), therefore $C_{\text{S1}} = 2$, while Sentinel-2 modalities contain up to 13 spectral bands depending on level and resolution. 
These modalities are treated as complementary views of the same location, acting as natural augmentations, which support our training strategy and encourage learning modality-invariant representations.

To provide semantic grounding, each sample \( s \) is assigned a high-level land use and land cover (LULC) category \( y^{(s)} \in \{1, \dots, Y\} \), derived from the ESA WorldCover product.
These categories reflect coarse semantic classes at a global scale and are used to compute class-frequency-aware statistics for balanced representation learning.

\noindent\textbf{Vision Transformer Model:}
\label{subsec:vit}

ViTs adapt the transformer architecture to visual data by treating an image as a sequence of patch tokens instead of a dense pixel grid. 
A typical ViT consists of two main components: a patch embedding module and a transformer encoder.
Given an input image \( \bm{x} \in \mathbb{R}^{H \times W \times C} \), the \textbf{patch embedding} layer \( f_{\theta} \) divides the image into \( N \) non-overlapping patches of size \( P \times P \), and projects each patch into a \( d \) dimensional embedding:
\[
\{\bm{z}_i\}_{i=1}^{N} = f_{\theta}(\bm{x}), \qquad \bm{z}_i \in \mathbb{R}^d.
\]
This projection is typically implemented using a convolutional layer with kernel size and stride equal to the patch size \( P \), parameterized by weights \( \mathbf{W}_\theta \in \mathbb{R}^{d \times C \times P \times P} \).
To encode spatial information, the transformer encoder augments each patch token \( \bm{z}_i \) with a positional vector. A learnable class token \( \bm{z}_{\text{cls}} \) is added to the sequence, which yields the full input:
\[
\bm{Z} = \bigl[\,\bm{z}_{\mathrm{cls}};\; \{\bm{z}_i + \text{pos}_i\}_{i=1}^{N}\,\bigr].
\]
The token sequence \( \bm{Z} \) is processed by a stack of \( L \) transformer layers, denoted \( \text{Enc}_\phi \). 
For classification tasks, only the final class token \( \hat{\bm{z}}_{\mathrm{cls}} \) is forwarded to a prediction head. 

\noindent\textbf{Modality-Specific Patch Embedding:}
\label{subsec:modality}

Standard patch embedding layers in ViTs are typically implemented using a shared convolutional projection across all inputs, making it unsuitable for multi-modal remote sensing data. 

To better handle this heterogeneity, we adopt a modality-specific patch embedding strategy. 
For each modality \( m \in \mathcal{M} \), we define an embedding function \( f_\theta^{m} \) that maps the input image \( \bm{x}^{m} \in \mathbb{R}^{H \times W \times C_m} \) to a sequence of patch tokens \( \bm{Z}^{m} \in \mathbb{R}^{N_m \times D} \), where \( C_m \) is the number of channels and \( N_m \) is the number of patches. Each \( f_\theta^{m} \) is parameterized independently to account for modality-specific dynamics.
We associate each modality with a learnable embedding vector \( \bm{e}^{m} \in \mathbb{R}^{D} \). This vector is added to every token from that modality via broadcasting:
\[
\tilde{\bm{Z}}^{m} = \bm{Z}^{m} + \mathbf{1}_{N_m} \cdot (\bm{e}^{m})^\top,
\]
where \( \mathbf{1}_{N_m} \in \mathbb{R}^{N_m \times 1} \) is a vector of ones. This allows the model to distinguish between modalities while preserving local spatial and spectral features.
Finally, to enable shared processing in the Transformer encoder, the enriched tokens \( \tilde{\bm{Z}}^{m} \) are linearly projected into a common latent space of dimension \( d \) using a shared projection \( \psi : \mathbb{R}^D \rightarrow \mathbb{R}^d \):
\[
\bm{Z}^{(m)} = \psi(\tilde{\bm{Z}}^{m}) \in \mathbb{R}^{N_m \times d}.
\]
This operation aligns all modality-specific token sequences in a unified representation space, allowing the encoder to process them jointly.

\noindent\textbf{Modality Augmentation and Cross-Attention Fusion:}
\label{subsec:modality_fusion}
Remote sensing observations of a single location are often captured using multiple sensors, each providing a unique spectral or radiometric perspective. 
Instead of treating these modalities as independent inputs, we interpret them as complementary views of the same scene. 
This allows us to use modality diversity as a form of natural augmentation, enabling the model to learn sensor-invariant representations.
In our setup, each spatial sample \( s \) from the Major-TOM dataset \cite{francis2024major} is observed via a fixed set of modalities. During pretraining, we independently assign modalities to the student and teacher networks via stochastic selection (threshold = 0.5), ensuring cross-modal supervision. E.g., the teacher may observe a global crop from Sentinel-1, while the student receives local views from Sentinel-2 L2A. This modality augmentation strategy encourages the model to align features across sensors, improving robustness to sensor-specific artifacts.
We consider two cases based on the number of selected modalities: 

\noindent\textbf{1) Single-Modality Views:}  
If only one modality is selected, the input is passed through the corresponding modality-specific patch embedding layer followed by the shared transformer encoder. This follows the standard ViT pipeline but uses modality-aware embeddings to handle spectral channel differences.
\textbf{2) Multi-Modality Fusion via Cross-Attention:}  
When multiple modalities are selected, we activate a modality fusion module based on cross-attention. For each selected modality \( m \in \mathcal{M} \), we obtain a patch token sequence \( \bm{Z}^{(m)} \in \mathbb{R}^{N \times D} \), where \( N \) is the number of spatial positions. These are stacked into a tensor \( \bm{Z}_{\text{all}} \in \mathbb{R}^{N \times M \times D} \), aligning spatial positions across modalities.

For each position \( n = 1, \dots, N \), we define shared learnable queries \( \bm{q} \in \mathbb{R}^{N_q \times D} \), which attend to modality-specific keys \( \bm{K}_n \in \mathbb{R}^{M \times D} \) and values \( \bm{V}_n \in \mathbb{R}^{M \times D} \), yielding \( N_q \) intermediate outputs:
\[
\bm{z}'_n = \mathtt{MultiHeadAttention}(\bm{q}, \bm{K}_n, \bm{V}_n) \in \mathbb{R}^{N_q \times D}.
\]
To aggregate them, we compute a learned weighted mean using softmax-normalized attention scores:
\[
\bm{w} = \mathtt{Softmax}( \bm{z}'_n \cdot \bm{p}_r ), \quad \bm{z}^{\text{fused}}_n = \sum_{j=1}^{N_q} w_j \bm{z}'_n[j],
\]
where \( \bm{p}_r \in \mathbb{R}^{D \times 1} \) is a learnable projection for scoring the query outputs. This results in a fused token \( \bm{z}^{\text{fused}}_n \in \mathbb{R}^D \). The final sequence \( \bm{Z}_{\text{fused}} \in \mathbb{R}^{N \times D} \) is then passed to the shared encoder.
This cross-attention fusion allows the model to dynamically weigh the modality contributions at each spatial location, capturing diverse information while maintaining spatial coherence.

\subsection{Pretraining}
\label{subsec:pretraining}
Our pretraining strategy builds on the DINO framework, which performs self-supervised learning.
It operates using a teacher-student setup, where both networks share the same ViT backbone and a lightweight three-layer projection head. Let \( g_{\theta_s} \) and \( g_{\theta_t} \) denote the student and teacher networks, respectively. While the student is trained using gradient-based optimization, the teacher is updated using EMA of the student's weights:
\begin{equation}
\theta_t \;\leftarrow\;
\lambda_e\,\theta_t + (1-\lambda_e)\,\theta_s,
\quad
\lambda_e = 1 - (1-\lambda_0)\frac{1+\cos\left(\pi e /E\right)}{2},
\label{eq:ema}
\end{equation}
where \( e \) is the current epoch, \( E \) is the total number of training epochs, and \( \lambda_0 \in [0.996, 1) \) is the initial momentum coefficient. The cosine schedule gradually increases \( \lambda_e \), stabilizing the teacher updates as training progresses.
This EMA mechanism allows the teacher to serve as a temporally smoothed ensemble of past student states, yielding more stable and consistent targets. Fig. \ref{fig:architecture} shows an overview of TerraFM pre-training.

\noindent\textbf{Multi-Crop Learning:}
\label{subsec:multi_crop}
To enable scale-invariant and cross-view representation learning, we adopt a multi-crop strategy as used in DINO \cite{caron2021emerging}. For each input sample, we generate two high-resolution global crops \( \{\bm{x}_g^1, \bm{x}_g^2\}  \subset \mathcal{X}_g \) and \( J \) low-resolution local crops \( \{\bm{x}_\ell^j\}_{j=1}^J \subset \mathcal{X}_\ell \). 
The teacher network processes only the global crops, while the student receives both global and local views. Each network produces a \( K \)-dim output which is temperature-scaled and normalized via the softmax function:
\[
Q_s(\bm{x})^{(i)} = \frac{\exp(g_{\theta_s}(\bm{x})^{(i)}/\tau_s)}{\sum_{k=1}^K \exp(g_{\theta_s}(\bm{x})^{(k)}/\tau_s)}, \quad
Q_t(\bm{x})^{(i)} = \frac{\exp((g_{\theta_t}(\bm{x})^{(i)} - c^{(i)})/\tau_t)}{\sum_{k=1}^K \exp((g_{\theta_t}(\bm{x})^{(k)} - c^{(k)})/\tau_t)},
\]
where \( \tau_s \) and \( \tau_t \) are temperature parameters that control output sharpness, and \( c \) is a centering term representing the running mean of teacher logits, used to stabilize training and avoid representation collapse.
The centering term is updated using exponential moving average over the teacher outputs:
\[
c \leftarrow \beta c + (1 - \beta) \cdot \frac{1}{B} \sum_{i=1}^{B} g_{\theta_t}(x_i),
\]
where \( \beta \in [0.9, 0.999] \) controls the momentum, and \( B \) is the batch size.
The overall loss encourages consistency between teacher and student predictions across all distinct view pairs:
\[
 \sum_{\bm{x} \in \mathcal{X}_g} \sum_{\substack{\bm{x}' \in \mathcal{X}; \bm{x}' \ne \bm{x}}} \mathcal{L}_{\text{CE}}\left(Q_t(\bm{x}), Q_s(\bm{x}')\right),
\]
where \( \mathcal{X} = \mathcal{X}_g \cup \mathcal{X}_\ell \), and \( \mathcal{L}_{\text{CE}}(\cdot, \cdot) \) denotes the cross-entropy loss. This loss formulation requires the student to produce consistent representations in all views.

\noindent\textbf{Dual Centering for Long-Tailed Distributions:}
\label{subsec:dual_centering}
Remote sensing datasets often exhibit long-tailed distributions of LULC classes, with frequent categories such as Forest dominating, while classes like Urban or Bare Land remain underrepresented as shown in Fig.~\ref{fig:map_and_pie}.
This imbalance persists even after subsampling and poses challenges for representation learning. Standard self-supervised approaches like DINO \cite{caron2021emerging} apply a single global centering term to stabilize training and avoid representation collapse, but they do not account for semantic imbalance in the data.
To address this, we propose a dual-centering scheme that combines global statistics with class-frequency-aware regularization. 
In addition to the standard global center vector \( c \), we introduce a secondary center \( c_h \), computed from a subset of samples belonging to high-frequency LULC classes, such as tree cover, grassland, and open seas, based on dataset-level statistics.
Given a batch of teacher logits \( g_{\theta_t}(x) \), the adjusted logits for training are computed as:
\[
\hat{g}(\bm{x}) = g_{\theta_t}(\bm{x}) - \alpha \cdot c - (1 - \alpha) \cdot c_h,
\]
where \( \alpha \in [0,1] \) balances the contribution of the global and frequency-aware centers. The vector \( c_h \) is updated via exponential moving average using only frequent-class samples within each batch.
This dual-centering mechanism serves two key purposes: \textbf{(i)} it preserves the stability benefits of global centering as in DINO, and \textbf{(ii)} it introduces a soft rebalancing bias that counteracts the overrepresentation of dominant classes in the feature space. In ablations (Table \ref{tab:abl-GEO-Bench}), this adjustment leads to more balanced representation learning and improved downstream performance, particularly for underrepresented LULC categories.

\label{subsec:downstream}

%% file: sec/02_t_dataset.tex
\section{Pretraining Data Sampling}

\begin{figure*}[t]
  \centering
  \includegraphics[width=0.55\textwidth]{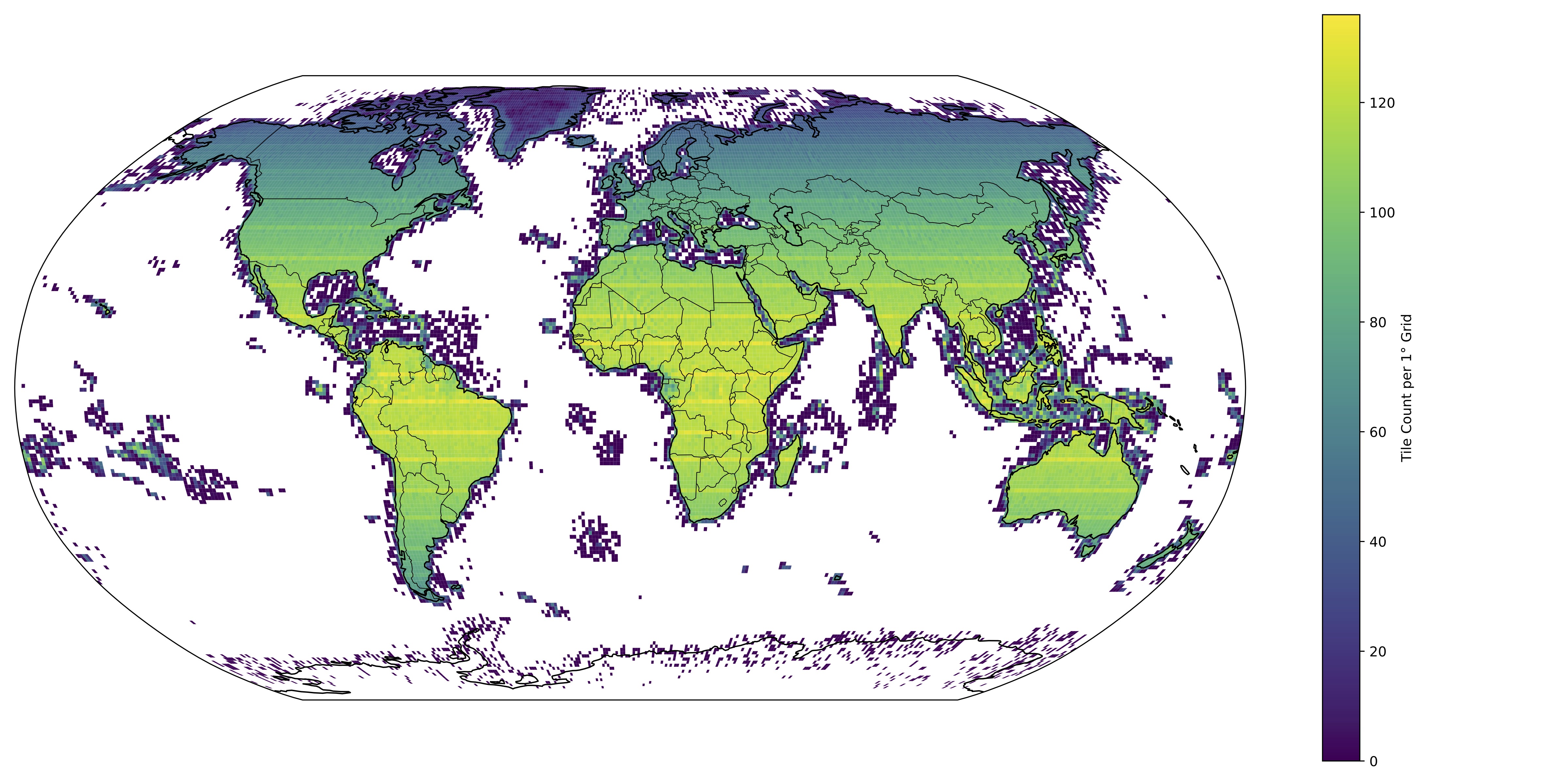}\hfill
  \includegraphics[width=0.45\textwidth]{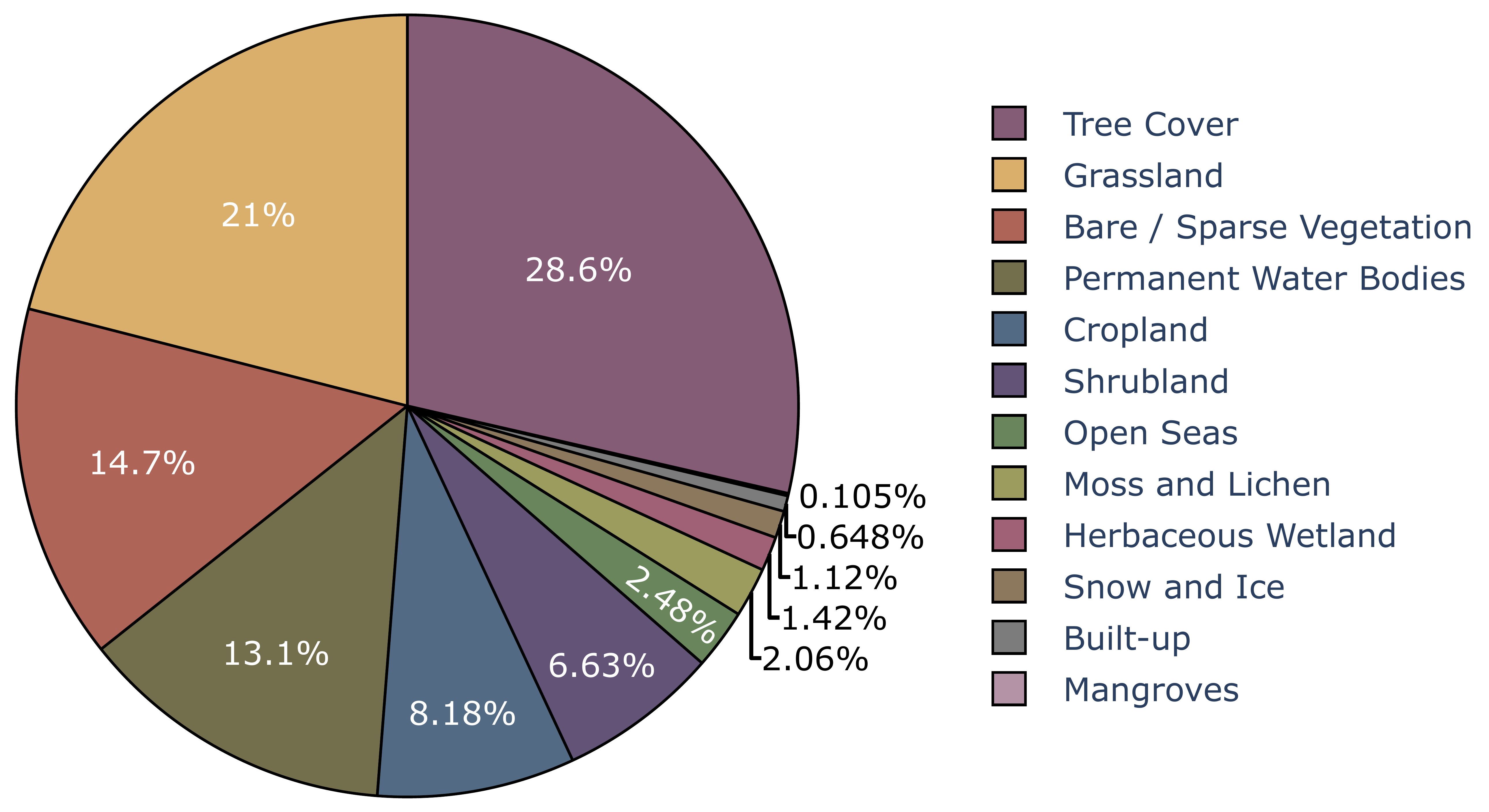}
  \caption{\emph{Left:} Global spatial distribution of the Major-TOM training subset. Each square shows a $1^\circ\times1^\circ$ cell, colored by the number of 10.68 km × 10.68 km tiles it contains.  
  \emph{Right:} Land-use/land-cover (LULC) breakdown across the same training tiles. A number of semantically important classes (e.g., builtup, mangroves, ice) remain underrepresented due to skewed data distribution.}
  \label{fig:map_and_pie}

\end{figure*}

We utilize the Major-TOM dataset~\cite{francis2024major} as our primary EO source for pretraining. 
It contains 2.24 million globally distributed grid cells, each spanning approximately 10.68 km $\times$ 10.68 km ($\approx$114 km$^2$), and provides tri-modal, co-registered imagery from Sentinel-2 Level-1C, Sentinel-2 Level-2A, and Sentinel-1 RTC. 
Major-TOM stands out as one of the few publicly available datasets offering dense multi-modal coverage at a global scale. 
However, over one-third of its samples lie outside a 10 km terrestrial buffer, often within the Open Oceans class~\cite{zanaga2022esa}, limiting their relevance for land-centric tasks. 
Motivated by insights from \cite{roscher2024better}, which emphasize the importance of semantically rich samples, and \cite{wang2024skyscript}, which highlight the utility of structural priors, we applied a principled filtering strategy. 
Specifically, we removed 98\% of ocean-classified tiles (retaining 2\% to preserve marine representation) and sampled the terrestrial subset using global distributional priors across land cover \cite{zanaga2022esa}, climate zones \cite{beck2018present}, and ESRI world regions \cite{esri_world_regions_2025}.
This approach emphasizes meaningful land regions with ecological variety. Fig.~\ref{fig:map_and_pie} shows the global coverage of the tiles.

For pretraining, we curated a filtered subset of over 1.5 million grid cells with consistent coverage across all three modalities (S1, S2-L1C, and S2-L2A). (Need to highlight the disk space usage, it takes around  76 terabytes to store 1.5M samples in uint16). 
Each 10.68 km $\times$ 10.68 km grid cell was divided into four non-overlapping tiles of 534 $\times$ 534 pixels, resulting in more than 6 million tiles per modality. 
In total, this yielded 18.7 million modality-specific training tiles.
During training, modalities were stochastically sampled and treated as natural augmentations to promote sensor-invariant representation learning.
To mitigate spatial sampling bias and support semantically-aware learning, we enriched each grid cell with metadata from the ESRI World Regions dataset~\cite{esri_world_regions_2025}.

%% file: sec/04_experiments.tex
\section{Experiments and Results}

\subsection{Pretraining Implementation Details}
We pretrain our TerraFM with a $16 \times 16$ patch resolution and an input size of $224 \times 224$. The training dataset comprises around 1.53 million multi-modal samples, from which we define a virtual epoch of 300K samples to ensure frequent parameter updates and improved memory efficiency. TerraFM-B is trained for 150 epochs and TerraFM-L is trained for 200 epochs with a linear warmup over the first 30 epochs.
Models are trained on 64 GPUs and the TerraFM-B training takes 92 hours with a batch size of 1024 where as the TerraFM-L use a  batch size of 2048 and training time is 183 hours. The learning rate is linearly scaled with batch size, initialized as $\text{lr} = 0.0001 \times \text{batch\_size}/256$. Following DINO-style pretraining, we disable batch normalization in the projection head and freeze the last layer of the student for the initial 3 epochs to stabilize early training. Following DINO \cite{caron2021emerging}, we use two global crops with scale sampled from $[0.25, 1.0]$ and six local crops from $[0.05, 0.25]$. The output dimensionality is set to $K=65{,}536$, with a teacher temperature schedule linearly increasing from 0.04 to 0.06 over the first 50 epochs. The momentum parameter for the teacher network follows a cosine schedule, starting from 0.996. A drop path rate of 0.1 is applied to regularize training. We set \( N_q = 5 \) and \( \alpha = 0.8 \) during pre-training.
\subsection{Evaluation Implementation Details}
\noindent\textbf{Linear Probing Evaluation: }
To evaluate the quality of learned representations, we follow a linear probing protocol of DINOv2\cite{oquab2023dinov2} that follows with a lightweight grid search over three key hyperparameters: (i) the learning rate, (ii) the number of transformer layers from which features are extracted, and (iii) whether to use only the [CLS] token or to concatenate it with the average-pooled patch tokens. 
We train the linear classifier using stochastic gradient descent (SGD) for 50 epochs. The training data is augmented using random resized cropping. Specifically, we sweep the learning rate over the set 
$\{10^{-5},\, 2 \times 10^{-5},\, 5 \times 10^{-5},\, 10^{-4},\, 2 \times 10^{-4},\, 5 \times 10^{-4},\, 10^{-3},\, 2 \times 10^{-3},\, 5 \times 10^{-3},\, 10^{-2},\, 2 \times 10^{-2},\, 5 \times 10^{-2},\,  1\}
$
Importantly, this search is computationally efficient: features from the frozen backbone are computed once per image using a single forward pass and reused across all configurations, since each linear head only requires a simple forward pass. For each configuration, we evaluate the classifier on the validation set and report the test accuracy achieved by the best validation configuration.
\noindent\textbf{UperNet Probing Evaluation: }
For UperNet \cite{upernet}  Probing evaluation, we freeze the pretrained backbone and attach \texttt{UPerNet} decoder head. Specifically, we use a \texttt{Feature2Pyramid} module as the neck, followed by a \texttt{UPerNet} decoder and an auxiliary \texttt{FCNHead}.
We train only the segmentation heads using the AdamW optimizer for 50 epochs without learning rate warm-up. We conduct a grid search over base learning rates
\(
\{10^{-1}, 10^{-2}, 10^{-3}, 10^{-4}, 10^{-5}, 10^{-6}\}.
\) and batch size set \(
\{16, 32, 64\}.
\)
\noindent\textbf{k-NN Evaluation: }
To assess the quality of the learned representations without any finetuning, we apply non-parametric classification using a $k$-nearest neighbors (k-NN) classifier on the frozen features. 
In addition to sweeping over $k \in {3, 5, 7, 10, 15, 20, 30, 50, 100}$ using validation performance, we follow the same layer selection strategy as linear probing i.e evaluating features from the last 4 transformer layers.
This protocol does not require additional training or data augmentation, making it a lightweight and reliable indicator of raw feature quality in pretrained models.
\noindent\textbf{Finetuning Evaluation: }
For full-model finetuning, we unfreeze the backbone and jointly optimize it with the task-specific head. We perform a grid search over learning rates in the set and batch sizes. To stabilize training, we apply a reduced learning rate for the backbone, set to half of the main learning rate used for the head parameters. Once the best configuration is selected based on validation performance, we evaluate the finetuned model on the test set.

\subsection{Evaluating Downstream Tasks}
\noindent\textbf{Benchmarks:}
We evaluate our model on two comprehensive remote sensing benchmarks: \textbf{GEO-Bench} and \textbf{Copernicus-Bench}, both of which include diverse downstream tasks spanning multiple domains and modalities. See more details in suppl. material.

\begin{table}[t]
\caption{We evaluate image classification using k-nearest neighbors (kNN) and report Top-1 accuracy for all single-label tasks. For the multilabel BigEarthNet benchmark, we report the F1 score. Results other than Copernicus-FM and TerraFM are directly taken from \cite{tseng2025galileo}.}
\label{tab:knn-results}
\centering
\resizebox{\textwidth}{!}{%
\begin{tabular}{lccccccccc}
\toprule
\multicolumn{2}{c}{\multirow{2}{*}{}} & \multicolumn{2}{c}{m-EuroSat}  & \multicolumn{2}{c}{m-BigEarthNet}  & \multicolumn{2}{c}{m-So2Sat}  & \multicolumn{2}{c}{m-Brick-Kiln}  \\
\multicolumn{2}{c}{}                  & \multicolumn{2}{c}{Training \%} & \multicolumn{2}{c}{Training \%} & \multicolumn{2}{c}{Training \%} & \multicolumn{2}{c}{Training \%} \\
\midrule     

Model              & Backbone         & 100\%           & 1\%            & 100\%           & 1\%            & 100\%           & 1\%            & 100\%           & 1\%            \\
\midrule     

SatMAE             & ViT-Base         & 84.1           & 34.8          & 50.6           & 29.0          & 36.0           & 23.1          & 86.1           & 73.5          \\
SatMAE++           & ViT-Large        & 82.7           & 48.5          & 50.8           & 31.6          & 34.7           & 23.4          & 89.6           & 76.7          \\
CROMA              & ViT-Base         & 85.6           & 51.3          & 58.8           & 44.7          & 48.8           & 33.8          & 92.6           & \underline{85.1}          \\
SoftCon            & ViT-Small        & 89.8           & 27.2          & 64.7           & 43.3          & 51.1           & 31.4          & 89.2           & 77.8          \\
DOFA               & ViT-Base         & 82.8           & 49.6          & 49.4           & 29.9          & 41.4           & 29.4          & 88.3           & 78.3          \\
Satlas             & Swin-Tiny        & 81.7           & 35.8          & 51.9           & 29.6          & 36.6           & 27.1          & 88.2           & 73.0          \\
MMEarth            & CNN-atto         & 81.7           & 30.0          & 58.3           & 39.6          & 39.8           & 25.1          & 89.4           & 79.7          \\
DeCUR              & ViT-Small        & 89.0           & 46.6          & 63.8           & \underline{49.6}          & 45.8           & 30.9          & 83.7           & 74.2          \\
AnySat             & ViT-Base         & 82.2           & 47.1          & 54.9           & 33.7          & 39.8           & 29.0          & 85.3           & 72.0          \\
Galileo            & ViT-Base         & 93.0           & 56.6          & 59.0           & 36.5          & 54.8           & \textbf{43.2}          & 90.7           & 78.0   \\
Prithvi-2.0        & ViT-Large        & 80.2           & 48.0          & 49.4           & 28.8          & 29.5           & 26.1          & 87.9           & 80.6          \\

Copernicus-FM & ViT-Base & 76.0 &47.4 & 53.8&33.3 & 38.4&23.3 & \underline{93.0}&83.2 \\ \midrule
\multirow{2}{*}{{TerraFM}}  &  ViT-Base & \cellcolor{orange!15} \underline{94.2}& \cellcolor{orange!15} \underline{59.3} & \cellcolor{orange!15} \underline{68.7}& \cellcolor{orange!15} 49.4 & \cellcolor{orange!15} \underline{55.1}& \cellcolor{orange!15} \underline{41.6} & \cellcolor{orange!15} \textbf{94.5}& \cellcolor{orange!15} \textbf{85.6} \\
 &  ViT-Large & \cellcolor{orange!15} \textbf{95.1} & \cellcolor{orange!15} \textbf{62.1} & \cellcolor{orange!15} \textbf{69.4}& \cellcolor{orange!15} \textbf{50.6} & \cellcolor{orange!15} \textbf{55.9} & \cellcolor{orange!15} 41.1 & \cellcolor{orange!15} \underline{93.0} & \cellcolor{orange!15} 82.2 \\
\bottomrule
\end{tabular}}
\end{table}

\noindent\textbf{Discussion:}
We report KNN classification accuracy on four standard GEO-Bench classification tasks to evaluate the quality of learned representations in a training-free setting. 
As shown in Tab.~\ref{tab:knn-results}, TerraFM achieves the highest performance across three datasets, outperforming both modality-specific and multimodal foundation models. 
Notably, our model achieves 95.1\% on m-EuroSAT and 94.5\% on m-Brick-Kiln, highlighting the effectiveness of the learned representations on standard scene classification tasks. 
On other challenging tasks such as m-So2Sat and m-BigEarthNet, our model achieves leading performance (55.9\% and 69.4\%, respectively), outperforming Galileo~\cite{tseng2025galileo}, despite So2Sat having fewer channels than used during pretraining, highlighting the model’s robustness to missing modality information.
Compared to CROMA~\cite{fuller2023croma} and DeCUR~\cite{wang2024decoupling}, our gains suggest that contrastive alignment combined with cross-modal fusion enhances class separability. 
The results across tasks of varying difficulty indicate that our model learns robust and transferable representations that generalize well across different scenarios.

\noindent Further on GEO-Bench, for classification (with fine-tuning), TerraFM achieves the improvement on m-BigEarthNet (73.1\%) and m-EuroSat (98.6\%), and  the best-performing model on m-So2Sat (66.6\%). 
For segmentation (with linear probing), our TerraFM-L notably outperforms existing models on m-SA-Crop-Type (34.5\% mIoU) and m-Cashew-Plant (37.2\% mIoU). Tab.~\ref{tab:benchGeo} shows that TerraFM-B surpasses larger counterparts such as ViT-Large used in SatMAE++ and DOFA.

On the Copernicus-Bench \cite{wang2025towards} evaluation, our model consistently outperforms existing foundation models across tasks and modalities (Tab.~\ref{tab:benchmarkcop}). 
A comprehensive comparison of TerraFM with existing methods on Copernicus-Bench, using metrics like OA (Overall Accuracy), mAP (mean Average Precision), and mIoU (mean Intersection over Union). 
Notably, TerraFM consistently achieves the highest scores across most of the tasks and metrics. 
In particular, it achieves an OA of 99.1\% on EuroSAT-S2, an mAP of 84.4\% on BigEarthNet-S2, and an mIoU of 67.9\% on Cloud-S2.

\begin{table*}[ht]
\centering
\caption{Comparison of TerraFM with existing supervised and self-supervised methods on Copernicus-Bench. Metrics include OA (Overall Accuracy) for classification tasks, mAP (mean Average Precision) for multi-label classification, and mIoU (mean Intersection over Union) for segmentation.}
\label{tab:benchmarkcop}
\resizebox{\textwidth}{!}{
\begin{tabular}{@{}lcccccccc}
\toprule 
 & \textbf{Metric} & \textbf{Supervised} & \textbf{Random} & \textbf{SoftCon} & \textbf{CROMA} & \textbf{DOFA} & \textbf{Copernicus-FM} & \textbf{TerraFM}  \\ \midrule
Backbone & -- & ViT-B/16 & ViT-B/16 & ViT-B/14 & ViT-B/8 & ViT-B/16 & ViT-B/16 & ViT-B/16   \\
\midrule

Cloud-S2       & mIoU & 59.4 & 60.4 & 66.9          & 65.0          & 65.0 & 66.7 & \cellcolor{orange!15} \textbf{67.9}  \\
EuroSAT-S1     & OA   & 81.5 & 75.4 & 83.6          & 83.9          & 81.7 & 87.2 & \cellcolor{orange!15} \textbf{87.8} \\
EuroSAT-S2     & OA   & 97.6 & 92.5 & 96.7          & 97.0          & 97.2 & 97.9 & \cellcolor{orange!15} \textbf{99.1} \\
BigEarthNet-S1 & mAP  & 70.6 & 63.8 & \textbf{78.7} & 70.8          & 70.5 & 77.9 & \cellcolor{orange!15} 76.9         \\
BigEarthNet-S2 & mAP  & 80.1 & 71.6 & 83.6          & 76.4          & 75.5 & 79.0 & \cellcolor{orange!15} \textbf{84.4}  \\
DFC2020-S1     & mIoU & 50.8 & 45.4 & 52.8          & 52.7          & 49.7 & 52.4 & \cellcolor{orange!15} \textbf{55.4}  \\
DFC2020-S2     & mIoU & 66.2 & 62.3 & 64.1          & \textbf{66.5} & 61.8 & 64.5 & \cellcolor{orange!15} 63.8        \\
LCZ-S2         & OA   & 85.3 & 77.4 & 83.6          & 84.1          & 83.0 & 84.4 & \cellcolor{orange!15} \textbf{87.0}

\\ \bottomrule
\end{tabular}}
\end{table*}

\begin{table*}[!htp]
\caption{Performance comparison on GEO-Bench for both classification (Top-1 Accuracy), segmentation (mIoU), and F1 score (for m-BigEarthNet). TerraFM achieves state-of-the-art results across multiple datasets, outperforming previous FMs.}
\label{tab:benchGeo}
\centering
\renewcommand{\arraystretch}{1.2}
\resizebox{\textwidth}{!}{
\begin{tabular}{@{}l l c c c c c c c@{}}
\toprule
 & \multicolumn{6}{c}{\textbf{Classification}} & \multicolumn{2}{r}{\textbf{Segmentation}} \\ 
\cmidrule(lr){3-6} \cmidrule(l){7-8}
Method & Backbone & m-EuroSat & m-BigEarthNet & m-So2Sat & m-Brick-Kiln & m-Cashew-Plant & m-SA-Crop-Type \\
\midrule

SatMAE      & ViT-Large & 96.6 & 68.3 & 57.2 & 98.4 & 30.8 & 24.8 \\
SatMAE++    & ViT-Large & 96.5 & 67.9 & 56.0 & 98.6 & 29.6 & 25.7 \\
CROMA       & ViT-Large & 96.6 & 71.9 & 60.6 & 98.7 & 31.8 & 32.0 \\
SoftCon     & ViT-Base  & 97.5 & 70.3 & 61.7 & 98.7 & 29.6 & 30.8 \\
DOFA        & ViT-Large & 96.9 & 68.0 & 58.7 & 98.6 & 27.7 & 25.4 \\
Satlas      & Swin-Base & 97.5 & \underline{72.8} & 61.9 & 98.4 & 25.1 & 23.4 \\
MMEarth     & CNN-atto  & 95.7 & 70.0 & 57.2 & \underline{98.9} & 24.2 & 22.2 \\
DeCUR       & ViT-Small & 97.9 & 70.9 & 61.7 & 98.7 & 26.2 & 21.5 \\
Prithvi 2.0 & ViT-Large & 96.5 & 69.0 & 54.6 & 98.6 & 26.7 & 22.9 \\
AnySat      & ViT-Base  & 95.9 & 70.3 & 51.8 & 98.6 & 26.1 & 27.1 \\
Galileo     & ViT-Base  & 97.7 & 70.7 & 63.3 & 98.7 & 33.0 & 30.1 \\
\midrule

\multirow{2}{*}{TerraFM}  
    &  ViT-Base  & \cellcolor{orange!15} \underline{98.1} 
    & \cellcolor{orange!15} 72.6 
    & \cellcolor{orange!15} \underline{64.9} 
    & \cellcolor{orange!15} 98.7 
    & \cellcolor{orange!15} \underline{34.1}  
    & \cellcolor{orange!15} \underline{33.0} \\
 &  ViT-Large 
    & \cellcolor{orange!15} \textbf{98.6} 
    & \cellcolor{orange!15} \textbf{73.1} 
    & \cellcolor{orange!15} \textbf{66.6} 
    & \cellcolor{orange!15} \textbf{99.0} 
    & \cellcolor{orange!15} \textbf{37.2} 
    & \cellcolor{orange!15} \textbf{34.5} \\
\bottomrule
\end{tabular}}
\end{table*}

\subsection{Ablations and Analysis}
\noindent\textbf{Imapct of Components:}
Tab.~\ref{tab:abl-GEO-Bench} highlights the incremental benefits of each component in our framework. 
We train TerraFM-B for 150 epochs on a 200k-sample subset from our full training dataset.
The model was trained on a subset of the training data, and KNN classification accuracy is reported for each dataset. To measure the performance on segmentation task, we use uppernet probing on the m-Cashew-Plantation dataset from GeoBench. 
Adding modality as augmentation improves performance on m-EuroSat by +4.5 and m-BigEarthNet by +3.01. 
Incorporating fusion yields a large gain on m-Cashew-Plantation by +11.82, while dual centering provides further improvements: +3.44 on m-BigEarthNet, +7.2 on m-EuroSat, and +14.0 on m-Cashew-Plantation.

\begin{SCtable}[0.8][!htp]
\caption{\footnotesize Ablation of components: SS = Self-supervised contrastive learning, MAug = Modality Augmentation, Fus = Fusion, DC = Dual Centering. BEN = m-BigEarthNet, ES = m-EuroSat, CP = m-Cashew-Plant.}
\centering
\scriptsize
\resizebox{0.5\linewidth}{!}{
\begin{tabular}{cccc|rrr}
\toprule
\textbf{SS} & \textbf{MAug} & \textbf{Fus} & \textbf{DC} & \textbf{BEN} & \textbf{ES} & \textbf{CP} \\
\midrule
\checkmark & -- & -- & -- & 54.62 & 83.20 & 50.58 \\
\checkmark & \checkmark & -- & -- & 57.63 & 87.70 & 59.17 \\
\checkmark & \checkmark & \checkmark & -- & 57.74 & 88.50 & 62.40 \\
\checkmark & \checkmark & \checkmark & \checkmark & 58.06 & 90.40 & 64.58 \\
\bottomrule
\end{tabular}
}
\label{tab:abl-GEO-Bench}
\end{SCtable}

\begin{figure*}[h]
    \centering
    \begin{minipage}[t]{0.49\textwidth}
        \centering
    \includegraphics[width=.99\linewidth]{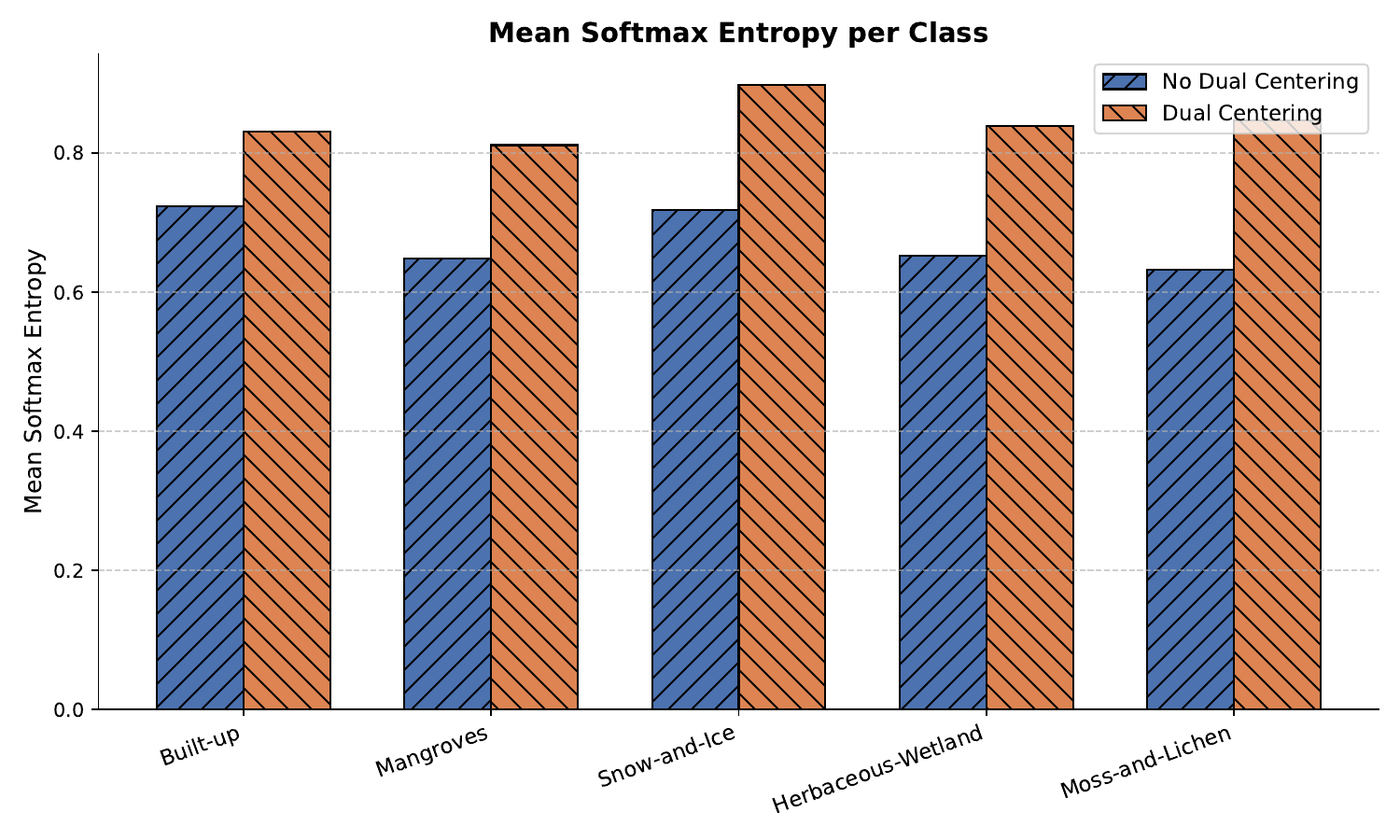}
    \captionsetup{justification=justified}
    \caption{Mean entropy per LULC class computed on 5k uniformly sampled training samples. Logits ($K{=}65,536$) are projected to a lower-dimensional space using a fixed random Gaussian matrix before computing entropy. The baseline model (No Dual-centering) exhibits lower entropy for most classes, showing overconfident predictions biased toward frequent-class prototypes. In contrast, our Dual-centering model yields higher entropy, suggesting reduced dominance of high-frequency prototypes, especially for rare classes (“Mangroves”, “Herbaceous-Wetland”).}
    \label{fig:rgb_vs_ms}
    \end{minipage}
    \hfill
    \begin{minipage}[t]{0.49\textwidth}
        \centering
    \includegraphics[width=\linewidth]{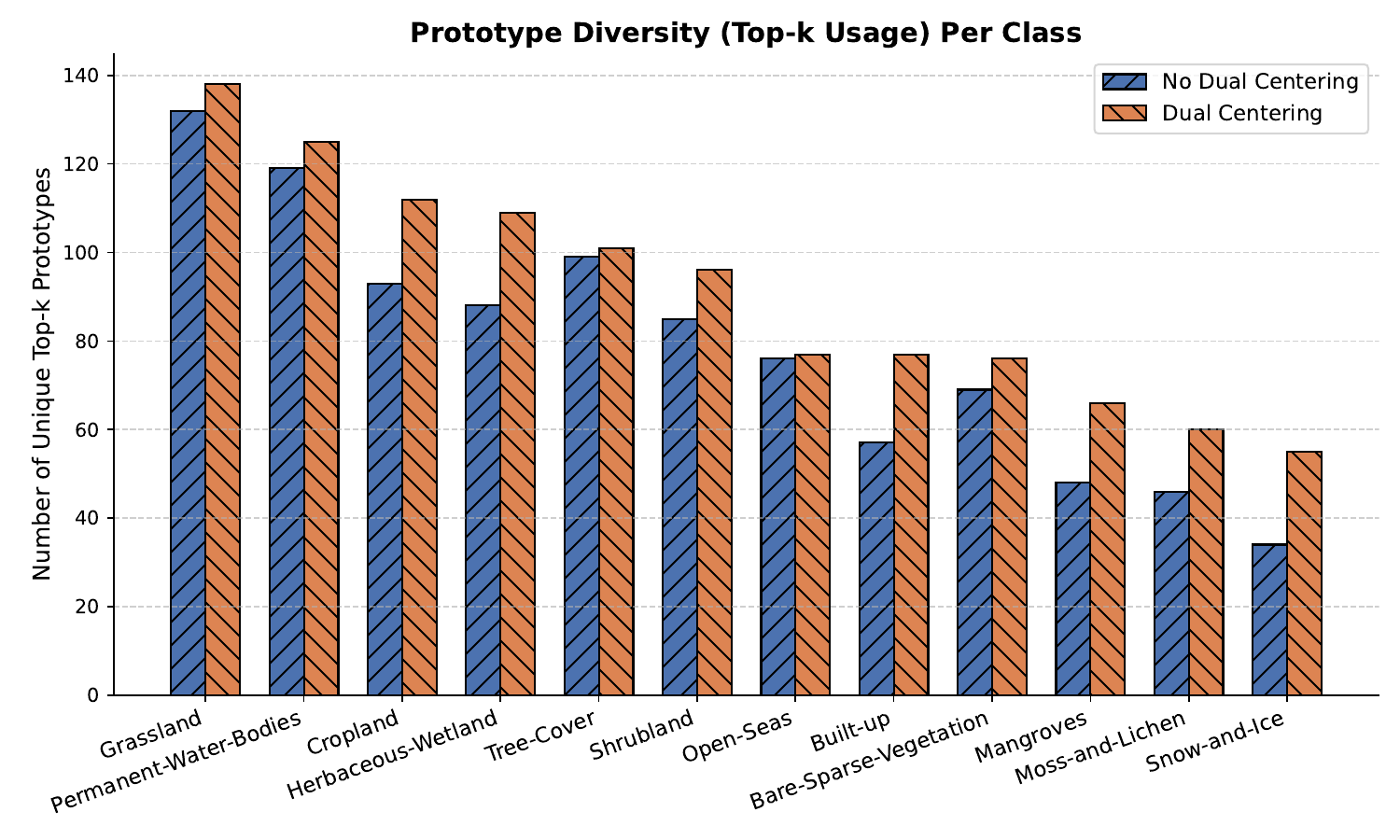}
    \captionsetup{justification=justified}
    \caption{Prototype diversity measured as the number of unique top-5 prototypes activated across 5k uniformly sampled training samples. Dual-centering leads to greater prototype diversity in tail classes such as “Mangroves”, “Herbaceous-Wetland”, and “Built-up”, suggesting more diverse representation learning. The baseline (No Dual-centering) tends to reuse a smaller subset of prototypes, especially for rare classes, reflecting over-reliance on dominant features from high-frequency categories.}
    \label{fig:bi_vs_mt}
    \end{minipage}
\end{figure*}

\noindent\textbf{Dual-centering Motivation and Visualization:}
Here, we discuss the impact of Dual-centering on class-wise prediction behavior and representation diversity. 
Fig.~\ref{fig:rgb_vs_ms} shows that models with Dual-centering exhibit higher softmax entropy across most classes, indicating more calibrated predictions, particularly benefiting rare classes like ``Mangroves”.
Fig.~\ref{fig:bi_vs_mt} reveals that Dual Centering significantly increases prototype diversity, i.e., the number of distinct top-5 features activated, especially for tail classes. 
This suggests that the model avoids collapsing onto frequent-class prototypes and learns more diverse, semantically rich representations. 
These results motivate Dual-centering as an effective strategy for reducing class imbalance effects in representation learning.

\begin{SCfigure*}[][h]
\caption{
Model accuracy vs. efficiency on m-EuroSat dataset.
Each point reports a backbone’s k-NN classification accuracy (y-axis) against its inference MACs in billions (log-scaled x-axis).
Marker shape and colour encode backbone size, as indicated by the inset legend.
We report here two variants of the TerraFM (Base, Large), which achieve the highest accuracy while maintaining moderate computational cost relative to both lightweight and heavyweight baselines.}
    \centering
    \includegraphics[width=0.38\linewidth]{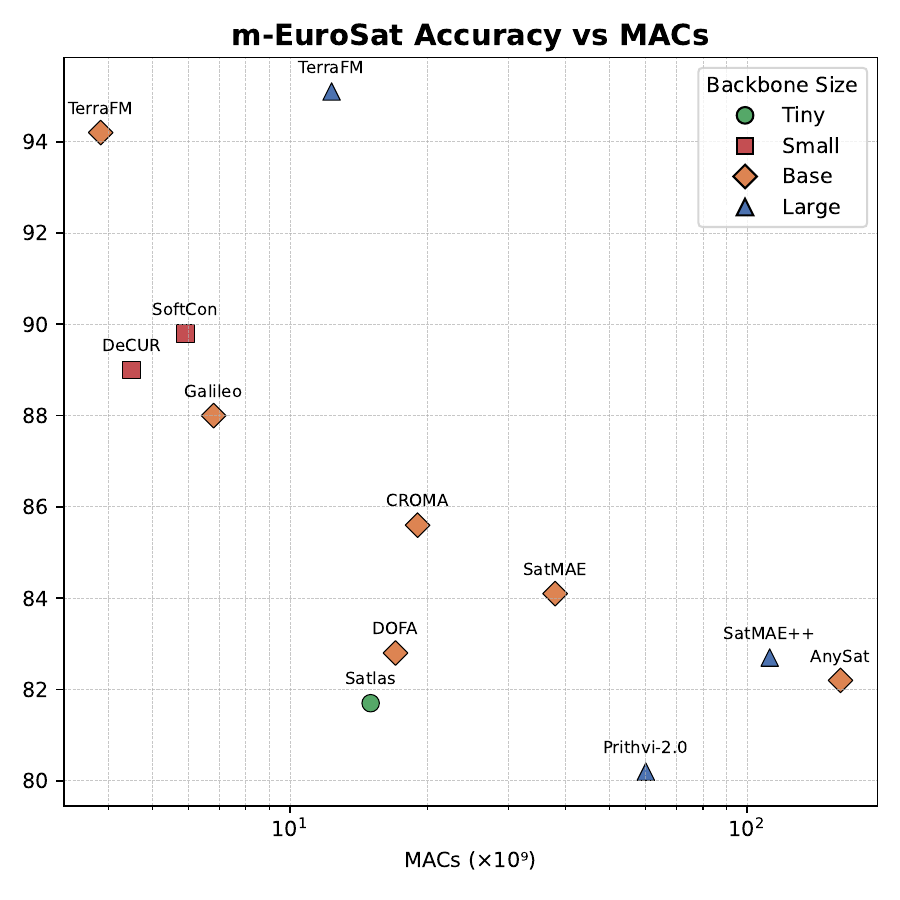}
    \label{fig:macs}
\end{SCfigure*}

\noindent\textbf{MACs-Performance Trade-Off:}
We evaluate the compute-efficiency trade-off of various remote sensing foundation models using Multiply-Accumulate operations (MACs) as a measure of inference cost. As shown in Fig.~\ref{fig:macs}, TerraFM achieves the highest accuracy on m-EuroSat while operating at significantly lower MACs compared to other large-scale models. 
This highlights the efficiency of our fusion design and pretraining strategy, demonstrating that strong performance can be achieved without excessive computational overhead. 
Models with higher MACs do not consistently translate to better accuracy, highlighting the importance of efficient and expressive architectures for scalable EO applications.

%% file: sec/05_supp.tex
This supplementary material presents additional experiments, analyses, and visualizations that complement the main paper. It includes detailed descriptions and experiments for our multimodal fusion strategies (\ref{sec:fusion}), 
and qualitative figures (\ref{sec:qual}). We also report GPU-hour comparisons with comparable methods (\ref{sec:gpuhours}), 
and visualize the land cover distribution of our dataset using global maps (\ref{sec:landcovermap}).


\section{Multi-Modal Fusion Strategies:}
\begin{figure*}[h]
    \centering
    \begin{minipage}[t]{\textwidth}
        \centering
    \includegraphics[width=\linewidth]{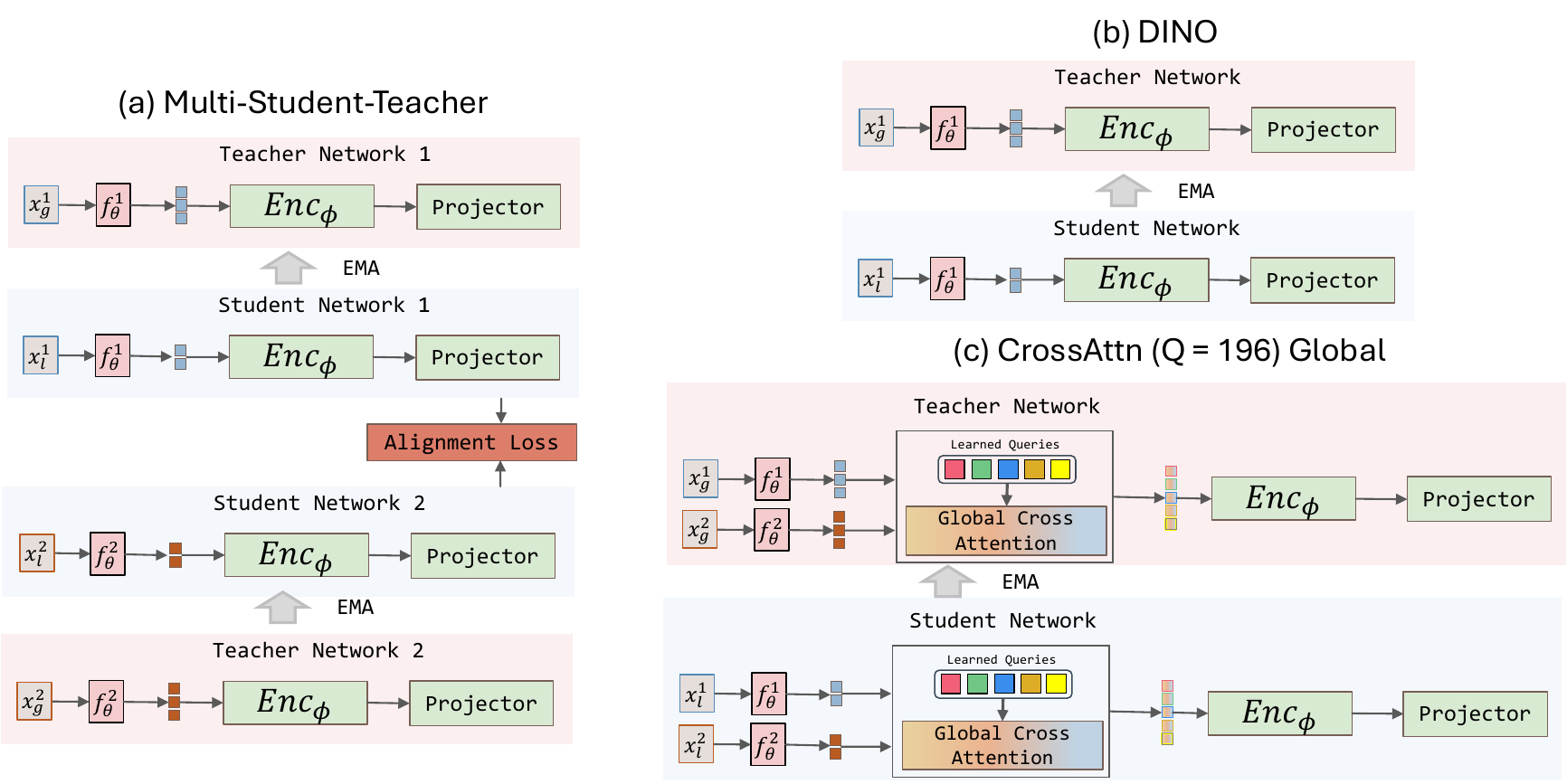}
    \captionsetup{justification=justified}
    \caption{Architectural overview of different fusion strategies: (a) Multi-Student-Teacher with alignment loss, (b) unimodal DINO baseline, and (c) CrossAttn (Q = 196) with global learned queries.}
    \label{fig:funsion_methods}
    \end{minipage}
    \hfill
    
\end{figure*}

\stepcounter{secnumber}
\label{sec:fusion}
We investigate various strategies for multi-modal fusion and report results in Table~\ref{tab:abl-fusion} on two benchmark datasets: m-BigEarthNet and m-EuroSat. As a baseline, we evaluate standard DINO training using only Sentinel-2 L2A input (\textit{DINO (S2-L2A)}), which learns unimodal representations.
To enable explicit modality-aware learning, we apply a \textit{Multi-Student-Teacher} approach where each modality has its own student and teacher networks, along with an alignment loss between student outputs to enforce cross-modal consistency. This yields consistent gains across both datasets. We also test a more expressive fusion approach, \textit{CrossAttn (Q = 196) Global}, where 196 learned queries (standard for 224×224 image inputs) attend globally to multi-modal tokens immediately after patch embedding. However, this method does not perform well, likely due to excessive parameterization and lack of inductive bias for spatial alignment. Figure~\ref{fig:funsion_methods} visually summarizes key fusion strategies evaluated in Table~\ref{tab:abl-fusion}, including (a) Multi-Student-Teacher, (b) unimodal DINO, and (c) CrossAttn (Q = 196) Global, highlighting their architectural differences and fusion mechanisms. Our proposed approach, \textit{TerraFM-B (Q = 1)}, treats a modality as an augmentation and performs fusion using a single learned spatial query per location. This lightweight attention mechanism yields the best performance among non-ensemble methods. To further analyze architectural choices, we test a variant, \textit{TerraFM-B (ViT PatchEmb)}, where the convolutional patch embedding is replaced by a ViT-S backbone purely for token extraction. While competitive, this setup slightly drops the performance due to increased model complexity and potential overfitting. Finally, our full model, \textit{TerraFM-B (Q = 5)}, employs multiple learned spatial queries to achieve richer fusion between modalities. It achieves the best overall performance, validating the scalability and effectiveness of our fusion design.

\begin{SCtable}[0.75][h]
\caption{Ablation study on multi-modal fusion strategies using k-NN evaluation. TerraFM-B with multiple spatial queries (Q = 5) achieves the best performance.}
\centering
\begin{tabular}{@{}lcc@{}}
\toprule
           & m-BigEarthNet             & m-EuroSat         \\   
\midrule
DINO (S2-L2A)          & 54.6 & 83.2 \\
Multi-Student-Teacher          & 55.8 & 87.8 \\
CrossAttn (Q = 196) Global & 52.0 & 77.1 \\
TerraFM-B (Q = 1) & 57.2 & 89.2 \\
TerraFM-B (ViT PatchEmb) & 56.9 & 87.2 \\
TerraFM-B (Q = 5) & \textbf{58.1} & \textbf{90.4} \\
\bottomrule
\end{tabular}
\label{tab:abl-fusion}
\end{SCtable}

\section{Qualitative:}
\stepcounter{secnumber}
\label{sec:qual}

\begin{figure*}[h]
    \centering
    \begin{minipage}[t]{\textwidth}
        \centering
    \includegraphics[width=\linewidth]{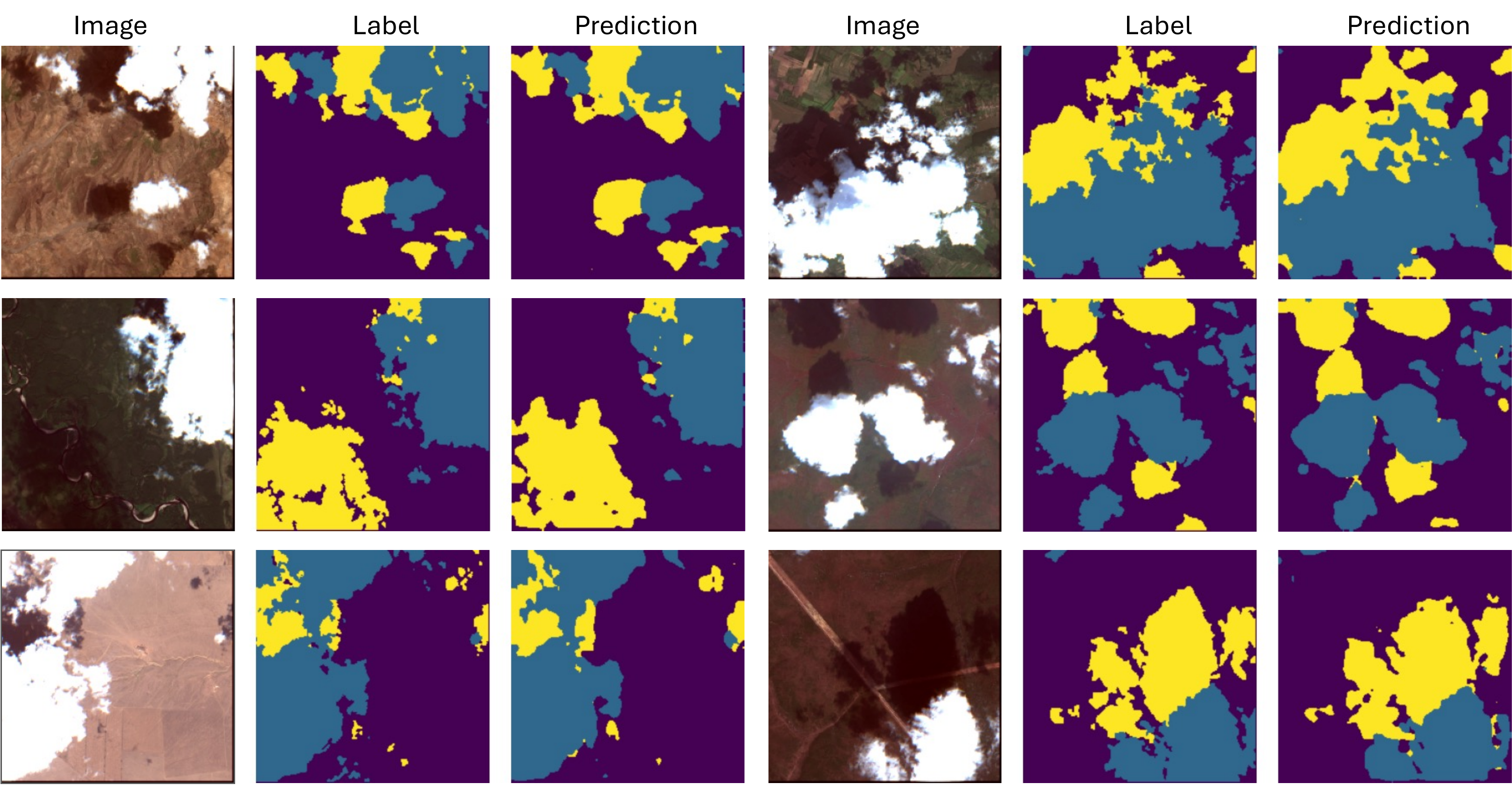}
    \captionsetup{justification=justified}
    \caption{Qualitative results for cloud and cloud shadow segmentation. Each triplet shows the input image (left), the ground truth mask (middle), and the TerraFM prediction (right).}
    \label{fig:qualitative}
    \end{minipage}
    \hfill
    
\end{figure*}

Fig.~\ref{fig:qualitative} illustrates qualitative results for the cloud and cloud shadow segmentation task. 
TerraFM accurately outlines both cloud and shadow regions, effectively distinguishing visually similar patterns while maintaining spatial coherence across varied scenes. 
These results demonstrate the model’s strong generalization ability under diverse and challenging atmospheric conditions.

\section{Landslide Detection}

\begin{figure*}[h]
    \centering
    \begin{minipage}[t]{\textwidth}
        \centering
    \includegraphics[width=\linewidth]{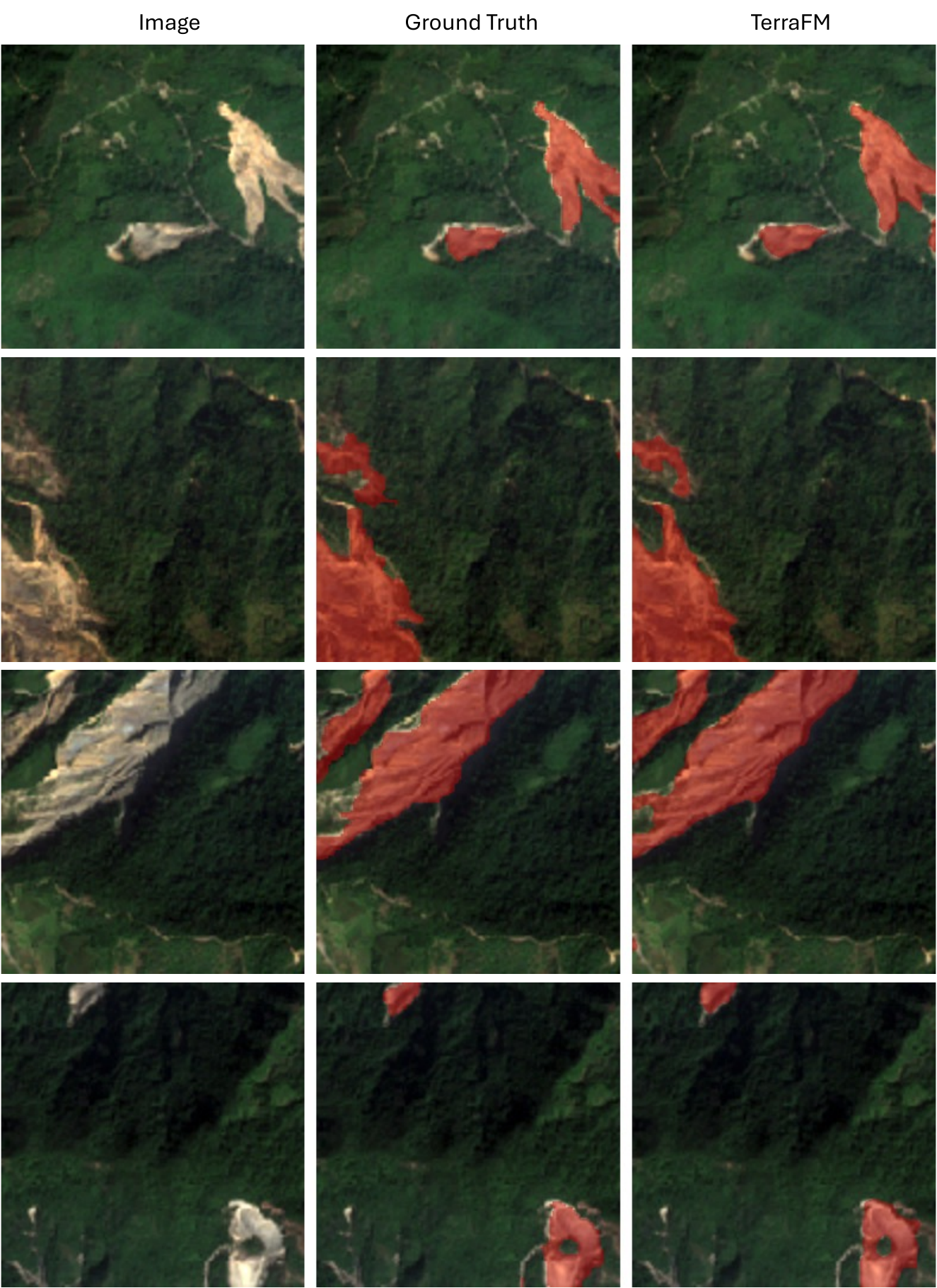}
    \captionsetup{justification=justified}
    \caption{Qualitative results for landslide segmentation. Each triplet shows the input image (left), the ground truth mask (middle), and the TerraFM prediction (right).}
    \label{fig:ls4s_qual}
    \end{minipage}
    \hfill
    
\end{figure*}

We evaluate landslide segmentation on the Landslide4Sense (L4S) benchmark, which provides segmentation labels for landslide and non-landslide regions across diverse mountainous areas using multi-source satellite data, including Sentinel-2 bands, DEM, and slope information. Our method, TerraFM, achieves strong performance with a mean IoU of 70.8 and a landslide IoU of 43.1, outperforming the Prithvi-EO-2.0 baseline (Table~\ref{tab:landslide}). Both TerraFM and Prithvi-EO-2.0 are trained using focal loss with a batch size of 16, Adam optimizer with a learning rate of $1\times10^{-4}$.
Figure~\ref{fig:ls4s_qual} shows qualitative results from TerraFM, illustrating predicted landslide masks alongside the ground truth.

\begin{SCtable}[0.75][h]
\caption{Landslide detection performance on the Landslide4Sense test set. Despite being significantly smaller (120M parameters vs.\ 300M for Prithvi-EO-2.0), TerraFM achieves higher overall segmentation performance, especially for landslide regions.}
\centering
\begin{tabular}{@{}lcc}
\toprule
           & mIoU             & IoU (Landslide)         \\   
\midrule
Prithvi-EO-2.0 (300M)  & 65.0 & 31.5 \\
TerraFM (120M)         & \cellcolor{orange!15} \textbf{70.8} & \cellcolor{orange!15} \textbf{43.1} \\
\bottomrule
\end{tabular}
\label{tab:landslide}
\end{SCtable}

\section{GPU Hour Comparison:}
\stepcounter{secnumber}
\label{sec:gpuhours}
Compared to Prithvi-2.0, which trains ViT-L (300M) model using up to 80 GPUs for 400 epochs, consuming approximately 21,000 GPU-hours~\cite{szwarcman2024prithvi}, our TerraFM (300M) achieves comparable scale using significantly fewer resources. Specifically, TerraFM is trained for 200 epochs on 64 GPUs, amounting to approximately 12,000 GPU-hours.



\section{Land Cover Distribution:}
\stepcounter{secnumber}
\label{sec:landcovermap}

\begin{figure}[h]
  \centering
    \includegraphics[width=1.0\linewidth]{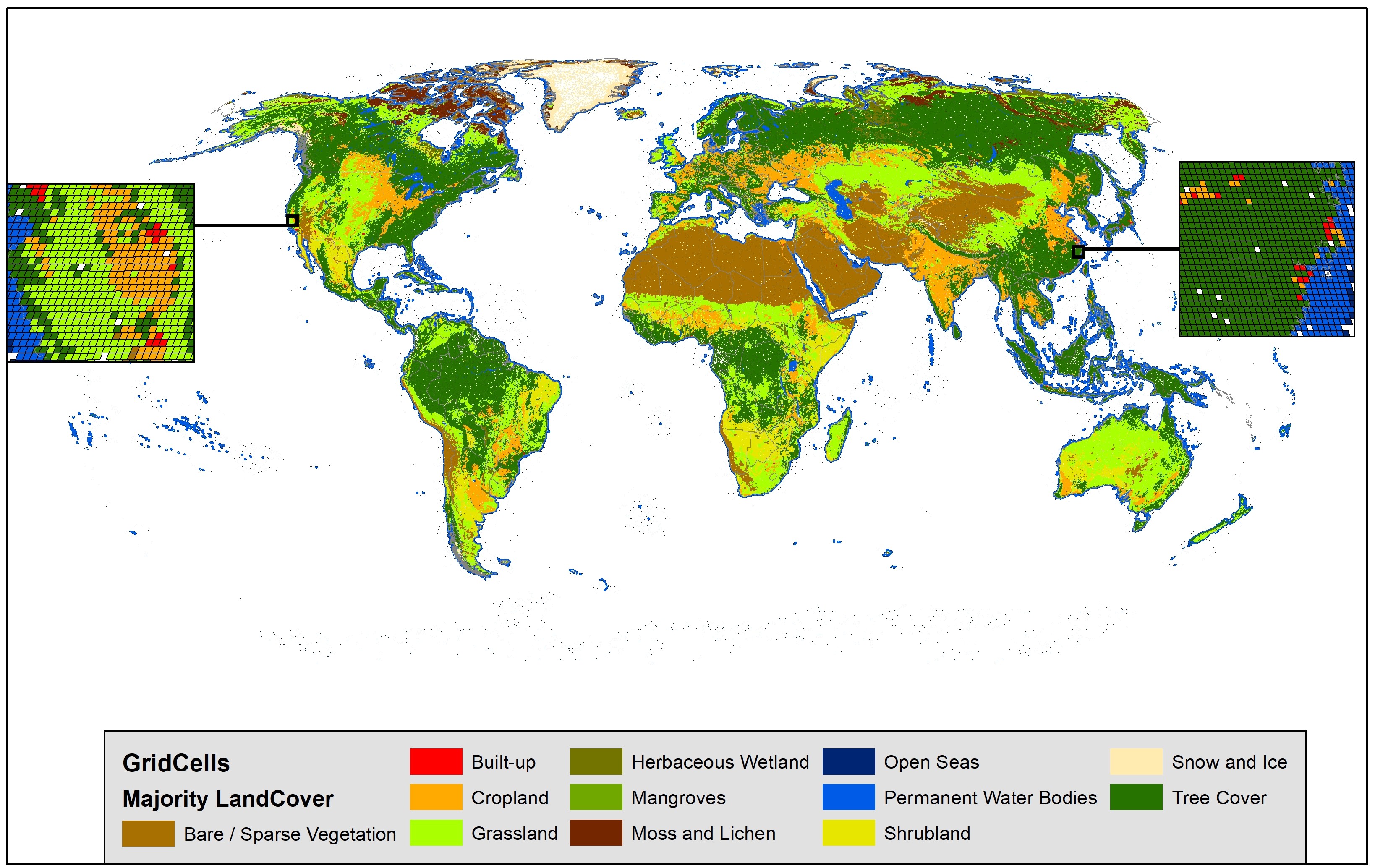}
    \caption{Global distribution of sampled training tiles by dominant land cover class, based on ESA WorldCover labels. Insets show detailed tile-level diversity, highlighting coverage across built-up, vegetation, and water classes.}
    \label{fig:landmap}
\end{figure}

Fig.~\ref{fig:landmap} illustrates the global spatial coverage of our pretraining data. 
The selected samples span diverse ecosystems, capturing a balanced mix of urban, vegetation, sea, and arid regions. 
The insets demonstrate fine-grained land cover variability, ensuring semantic richness across training tiles. 
This diverse geographic grounding plays a crucial role in enabling the generalization capabilities of TerraFM across regions and tasks.

